%% file: main.tex
\documentclass[runningheads]{llncs}

\usepackage{eccv}

\usepackage{eccvabbrv}

\usepackage{graphicx}
\usepackage{booktabs}

\usepackage[accsupp]{axessibility}  % Improves PDF readability for those with disabilities.

\usepackage{hyperref}

\usepackage{orcidlink}
\usepackage{makecell}
\usepackage{multirow}

\newcommand\nnfootnote[1]{%
  \begin{NoHyper}
  \renewcommand\thefootnote{}\footnote{#1}%
  \addtocounter{footnote}{-1}%
  \end{NoHyper}
}

\begin{document}

% ---------------------------------------------------------------
% TODO REVIEW: Replace with your title
\title{Defending Text-to-image Diffusion Models: Surprising Efficacy of Textual Perturbations Against Backdoor Attacks}

% TODO REVIEW: If the paper title is too long for the running head, you can set
% an abbreviated paper title here. If not, comment out.
\titlerunning{Surprising Efficacy of Textual Perturbations Against Backdoor Attacks}

% TODO FINAL: Replace with your author list. 
% Include the authors' OCRID for the camera-ready version, if at all possible.
\author{Oscar Chew\inst{1}$^*$\orcidlink{0009-0004-0330-7531}\and
Po-Yi Lu\inst{2}$^*$\orcidlink{0009-0009-8131-2666} \and
Jayden Lin\inst{3}\orcidlink{0009-0005-8090-7319} \and
Hsuan-Tien Lin\inst{2}\orcidlink{0000-0003-2968-0671}}

% TODO FINAL: Replace with an abbreviated list of authors.
\authorrunning{O. Chew et al.}
% First names are abbreviated in the running head.
% If there are more than two authors, 'et al.' is used.

% TODO FINAL: Replace with your institution list.
\institute{ASUS\and National Taiwan University\and University of Michigan \\
\email{oscar\_chew@asus.com \{d09944015,htlin\}@csie.ntu.edu.tw jaydelin@umich.edu}
}

\maketitle
\vspace{-1.5em}
\input{abstract}
\input{introduction}
\input{related_work}

\input{textual_perturbation}

\input{experiments}

\input{conclusion}
\input{acknowledgements}
% \clearpage\mbox{}Page \thepage\ of the manuscript.
% \clearpage\mbox{}Page \thepage\ of the manuscript.
% \clearpage\mbox{}Page \thepage\ of the manuscript.
% \clearpage\mbox{}Page \thepage\ of the manuscript.
% \clearpage\mbox{}Page \thepage\ of the manuscript. This is the last page.
% \par\vfill\par
% Now we have reached the maximum length of an ECCV \ECCVyear{} submission (excluding references).
% References should start immediately after the main text, but can continue past p.\ 14 if needed.
% \clearpage  % TODO REVIEW/FINAL: This \clearpage needs to be removed from both review and camera-ready versions.

% ---- Bibliography ----
%
% BibTeX users should specify bibliography style 'splncs04'.
% References will then be sorted and formatted in the correct style.
%
\bibliographystyle{splncs04}
\bibliography{main}
\appendix
\input{perturbation}
\input{training}
\input{rr-viz}
\input{cross-attention-viz}
\end{document}

%% file: abstract.tex
\begin{abstract}
Text-to-image diffusion models have been widely adopted in real-world applications due to their ability to generate realistic images from textual descriptions. However, recent studies have shown that these methods are vulnerable to backdoor attacks. Despite the significant threat posed by backdoor attacks on text-to-image diffusion models, countermeasures remain under-explored. In this paper, we address this research gap by demonstrating that state-of-the-art backdoor attacks against text-to-image diffusion models can be effectively mitigated by a surprisingly simple defense strategy—textual perturbation. Experiments show that textual perturbations are effective in defending against state-of-the-art backdoor attacks with minimal sacrifice to generation quality. We analyze the efficacy of textual perturbation from two angles: text embedding space and cross-attention maps. They further explain how backdoor attacks have compromised text-to-image diffusion models, providing insights for studying future attack and defense strategies. Our code is available at \url{https://github.com/oscarchew/t2i-backdoor-defense}.

% \htcomment{
% If I understand English correctly, every ... needs to use singular form? (I can be wrong)
% }
% \htcomment{
% This further reveals---not sure what "this" refers to. Maybe use "they" (i.e. the two angles)
% }
% \htcomment{
% maybe some change of words:
% They further explain how backdoor attacks have compromised text-to-image diffusion models, shedding lights on insights for studying future attack and defense strategies.
% }
% \htcomment{
% This sentence sounds weak. "Experiments show that textual perturbations are effective in mitigating backdoor attacks." Say a bit more on how effective it is, or maybe its effect across different t2i models
% }

% \htcomment{
% "Additionally, we unveil the changes in the text embedding space and the cross-attention maps in the presence of backdoor triggers to further explain the effectiveness of this approach." Maybe something more like "We analyze the reasons on the efficacy of the purturbation defense from two angles: the text embeddings, and the cross-attention maps, which reveals how the backdoor attack have broken different text-to-image models from different causes, shedding lights for studying future attacking and defense strategies."
% }

%\keywords{Generative AI \and Diffusion Model \and Backdoor Attack}
\vspace{-1em}
\end{abstract}

%% file: introduction.tex
\section{Introduction}
Text-to-image diffusion models \cite{ramesh21azero, nichol22glide, saharia22imagen} have significantly advanced the field of generative art, with Stable Diffusion \cite{Rombach_2022_CVPR} emerging as one of the leading approaches. Despite the tremendous success, the dark side of these models is often overlooked. These models, while powerful, are vulnerable to various security threats, including backdoor attacks. Such attacks can manipulate the output images in subtle yet malicious ways, posing significant risks to the integrity of the generated content \cite{Struppek_2023_ICCV, huang2024personalization, chou-etal-2023-villandiffusion}. Therefore, developing defense methods to mitigate backdoor attacks on text-to-image models is a critical research problem.
\nnfootnote{$^*$ Equal contribution}

%In contrast to the rapid development of backdoor attack methods, we observe that defense strategies for text-to-image diffusion models remain significantly under-studied. 
% Most existing defenses are designed for classification tasks \cite{gao2019strip, yang-etal-2021-rap,xue2023detecting,shi2023black}, leaving a gap in the context of text-to-image generation.
While defenses for classification tasks are well-studied\cite{gao2019strip, yang-etal-2021-rap,xue2023detecting,shi2023black}, defenses for text-to-image generation remain under-explored.
% \htcomment{
% While there are many defense methods for classification tasks~\cite{}, defenses for text-to-image generation are less explored.}
% \htcomment{
% These first few sentences seem redundant to the last sentence in the previous paragraph. Or at least it is not clear whether "robust methods" are the same as "defense strategies."
% }
Backdoor attacks generally work by injecting a text-based backdoor trigger. Hence, in this paper, we explore the idea of introducing perturbations into text inputs to disrupt these backdoor triggers.
% \htcomment{
% "explore the idea of" sounds jumpy---there is no clue on why such attacks might be effective for backdoor triggers. Maybe a bit more context on how the backdoor attack works before we "explore the idea"
% }
By applying semantic-preserving perturbations to the input text, we can disrupt the specific trigger patterns embedded in the text, thereby evading the backdoor attack with minimal sacrifice to the quality of the generated images. To justify the effectiveness of textual perturbation, we examine changes in both text embedding space and cross-attention maps under backdoor attacks. Our first key insight is that the injection of a backdoor trigger pushes it away from its initial neighbors in the text embedding space, suggesting these initial neighbors from textual perturbation could help evade backdoor attacks. Secondly, we find that perturbing the input text prevents the trigger token from hijacking the attention mechanism, thus avoiding the generation of malicious content.

%Perturbations have been identified as a promising strategy to mitigate backdoor attacks in the context of image classification, as discussed in \cite{gao2019strip,xue2023detecting}. 
Our analysis covers latest backdoor attacks and show that textual perturbation can mitigate these backdoor attacks effectively while maintaining the fidelity of the generated images. We summarize our key contributions as follows: \begin{itemize}
    \vspace{-0.5em}
    \item We design a simple yet effective textual perturbation strategy to mitigate state-of-the-art backdoor attacks against text-to-image diffusion models.
    \item We provide insights into how the text embedding space, as well as the cross-attention map, are altered in the presence of backdoor triggers.
    \item To the best of our knowledge, we are among the first to address backdoor attacks on text-to-image diffusion models.
    % \poy{As we know UFID and T2IShield, we are not the first one. How about `We are among the first teams to address backdoor attacks in text-to-image diffusion models. Moreover, we have gathered a broader range of attack methods and more comprehensive defenses.'}
    % \oscar{Isn't ``one of the best'' the same as ``among the first''? In general, I would like to claim T2IShield as contemporaneous and present a preliminary comparison to demonstrate some of our edges. (Over)-claiming ``more comprehensive defense'' as our contribution might set a very high expectation on the comparison.}
    
\end{itemize}

%% file: related_work.tex
\section{Related Work}
\subsection{Text-to-Image Diffusion Model}
Text-to-image diffusion models generate images by progressively refining noisy inputs through iterative processes guided by textual information. Stable Diffusion~\cite{Rombach_2022_CVPR}, as a notable example, leverages a pre-trained CLIP text encoder \cite{radford2021clip} to derive a conditioning vector from the input text. This conditioning vector plays a crucial role in enabling the model to generate images that accurately reflect the semantic content of the provided textual descriptions. 

% \poy{add personalization techniques for \cite{huang2024personalization}}
% Oscar reply: I will add the sentences below if space permits
%To accommodate novel, user-provided concepts, a line of research focuses on the personalization of text-to-image diffusion models \cite{Rinon_2023_ICLR, Ruiz_2023_CVPR, kumari2022customdiffusion}. For example, Textual Inversion \cite{Rinon_2023_ICLR} encode new concepts into pre-trained models by learning new tokens in the text embedding space with a few example images.

\subsection{Backdoor Attack against Text-to-Image Diffusion Models}
Struppek \etal \cite{Struppek_2023_ICCV} is the first to show that text-to-image diffusion models could be backdoored by manipulating the pre-trained text encoders. Their method, Rickrolling, uses a homoglyph (a visually similar non-Latin character) as a backdoor trigger.
VillanDiffusion~\cite{chou-etal-2023-villandiffusion} fine-tunes the U-Net component of diffusion models %with LoRA \cite{hu2022lora}
to inject backdoor triggers by manipulating the loss function. Huang \etal \cite{huang2024personalization} proposed that personalization techniques for diffusion models such as Textual Inversion \cite{Rinon_2023_ICLR} can be exploited to implant backdoor triggers by providing mismatched text-image pairs. Their potential countermeasures are believed to require human intervention or a copious amount of tests, according to \cite{Struppek_2023_ICCV, huang2024personalization}. 
% \htcomment{
% Also, consider adding some contents about the difficulty of the task
% }

\subsection{Backdoor Defense for Diffusion Models}
To the best of our knowledge, \cite{an2024elijah} is the only defense against backdoor attacks that have been published in a scientific venue. However, it is specifically tailored to the context of unconditional generation, whereas our work focuses on the setting of text-to-image generation. \cite{wang2024t2ishield} is a contemporaneous work addressing backdoor attacks on text-to-image diffusion models. While both our work and \cite{wang2024t2ishield} perform well in mitigating backdoor attacks, the insights offered by both works are complementary. \cite{wang2024t2ishield} discovers the ``Assimilation Phenomenon'' through the lens of cross-attention whereas our work provides a different view on the cross-attention maps and further sheds light on the changes in the text embedding space under backdoor attack. We will present a preliminary comparison with \cite{wang2024t2ishield} in \cref{sec:t2ishield-comparison} to demonstrate the edge of our approach.

%% file: textual_perturbation.tex
\section{Textual Perturbation as a Remedy}
Our proposed approach is a simple plug-and-play module that leverages textual perturbation to evade trigger tokens and thereby achieve enhanced security. The process is straightforward: before feeding the input text into CLIP text encoder, we transform the text using our proposed perturbations according to predetermined probabilities. The transformed sentence is then processed by the text encoder to obtain a conditioning vector, which is subsequently used by a U-Net to generate images. We consider the following semantic-preserving transformations as our textual perturbations. \Cref{tab:perturbation-examples} shows some examples of our textual perturbations. Details about the implementation can be found in \cref{app:perturbation}.

\begin{table}[bt]
    \caption{Examples of our perturbation strategies which aim to disrupt trigger tokens without affecting the original semantics}
    
    \label{tab:perturbation-examples}
    \centering
    \begin{tabular}{c|c|c}
      \toprule
       Perturbation strategy & Input & Output  \\
       \midrule
       Synonym replacement & beautiful \textcolor{red}{car} & beautiful \textcolor{red}{automobile}\\
       Translation & white \textcolor{red}{cat} & white \textcolor{red}{gato} \\
       Random character & \textcolor{red}{beautiful} car & \textcolor{red}{beautful} car \\
       Homoglyph replacement & h\textcolor{red}{\underline{o}}use & \textcolor{red}h{o}use\\
       \bottomrule
    \end{tabular}
\end{table}
% $\paragraph{Attack Scenario and Attach Capabilities}
% \label{sec:attack_scenario}
% \poy{(0) In politics \& business, the name of opponent can be used as a backdoor to discredit the opponent. In social network, the platform might be injected a backdoor to let audience see NSFW (1) Full control of the fine-tuning procedure. Neither access to control the systems after deployment.}$

\paragraph{Word-level Perturbation} This includes synonym replacement and translation. We randomly replace words with their synonyms based on the text embedding space \cite{mrksic-etal-2016-counter}. We leverage pre-trained models from OPUS-MT \cite{tiedemann2023democratizing, TiedemannThottingal:EAMT2020} to translate parts of the text from English to other languages, such as Spanish.

\paragraph{Character-level Perturbation}
This includes homoglyph replacement and random perturbation.
While Struppek \etal \cite{Struppek_2023_ICCV} claim that single non-Latin characters are not detectable by the naked eye, we argue that they can, in fact, be easily detected and handled by the system.
%\poy{(0) The `model deployer' appears somewhat abruptly in this paragraph. How about `While Struppek \etal~\cite{Struppek_2023_ICCV} claim that the single non-Latin characters are not detectable by the naked eye, these subtle homophones can still be systematically handled well.' (1) It recalls us to add \cref{sec:attack_scenario} to explain the attack scenario.}
Since the presence of non-Latin characters can often cause harm, we map non-Latin characters in sentences to visually similar Latin characters using a pre-defined dictionary.
We also perform additional random character deletion, swap, and insertion under constraints to perturb tokens without substantially impacting the original semantics.

%% file: experiments.tex
\section{Experiments}
\subsection{Experiment Setup}
\label{sec:exp-setup}
\paragraph{Models}
We consider latest backdoor attacks against text-to-image diffusion models, namely Rickrolling \cite{Struppek_2023_ICCV}, VillanDiffusion\cite{chou-etal-2023-villandiffusion} and Textual Inversion\cite{Rinon_2023_ICLR}.
We set the victim model to be Stable Diffusion v1.4. The training details as well as the hyperparameters are presented in \cref{app:training}.

\paragraph{Datasets} The datasets and triggers are adapted from the original implementation of each work. Specifically, the datasets used are MS COCO \cite{lin2014mscoco}, CelebA-Dialog \cite{jiang2021talk}, and four images of Chow Chow (a species of dog) for Rickrolling, VillanDiffusion\footnote{Chou \etal \cite{chou-etal-2023-villandiffusion} also adopt Pokemon Caption Dataset \cite{pinkney2022pokemon} in their experiments. However, the dataset is currently unavailable due to a DMCA takedown notice from The Pokémon Company International.}, and Textual Inversion respectively. Rickrolling associates U+0B20, U+0585 with ``A lightning strike'' and ``A blue boat on the water''. VillanDiffusion associates ``latte coffee'' and ``mignneko'' with an image of a cat. Finally, Textual Inversion associates ``beautiful car'' and ``[V]'' with the images of Chow Chow.

\paragraph{Metrics} We use Attack Success Rate (ASR) and Fréchet Inception Distance (FID) \cite{heusel2017fid} to evaluate the effectiveness of our method in preventing the generation of target images and assessing the fidelity of generated images for benign captions. ASR is defined as the rate at which generated images are classified as the class of the target image by a pre-trained CLIP model. FID measures the similarity between two sets of images by comparing the distributions of features extracted from a pre-trained network, thereby assessing the similarity between generated images and real images. Following the setting described by \cite{chou-etal-2023-villandiffusion}, we sample 3000 benign captions from CelebA-Dialog for the computation of FID.
% The datasets of the computation of ASR are the same as above.

\subsection{Qualitative Results}
First, we showcase how slight perturbations in the input text can mitigate backdoor attacks by reproducing backdoor attacks and then applying perturbations. \Cref{tab:qualitative} shows that every backdoor attack could be mitigated just by disrupting backdoor triggers. For instance,  although Textual Inversion ties ``beautiful car'' to the concept of Chow Chow, the prompt ``beautful car'' generates a photo of a car correctly; As for Rickrolling, it is straightforward that the generated images are faithful as the backdoor trigger no longer presents. Thus, it is evident that textual perturbations are effective against a wide variety of backdoor attacks.
%First, we reproduce three backdoor attacks with their pre-defined triggers corresponding to the target images or prompts. We then apply our proposed textual perturbation to the input texts and observe the changes in the input texts along with generated images.
\begin{table}[tb]
  \begin{center}
    \caption{Backdoor attacks are mitigated by slight textual perturbations.}
    \label{tab:qualitative}
    \resizebox{\linewidth}{!}{ %< auto-adjusts font size to fill line
      \input{tables/qualitative.tex}
    }
  \end{center}
\end{table}

\subsection{Quantitative Results}
\Cref{tab:effectiveness} shows that while Stable Diffusion is highly vulnerable to existing backdoor attacks, it can greatly benefit from incorporating simple textual perturbations. In many cases, the ASR decreases from 1 to 0, indicating an effective defense. Moreover, we observe a small decrease in FID, suggesting that the disruption to the semantics of the original text is within an acceptable range.
 % TODO show Cyrillic o and Greek o on LaTeX
\begin{table}[tb]
    \centering
    \caption{Effectiveness of textual perturbations against existing backdoor attacks}
    \small
    \begin{tabular}{lccccc}
    \toprule
        & & \multicolumn{2}{c}{No defense} & \multicolumn{2}{c}{Ours} \tabularnewline
        Attack method & Trigger & ASR ($\downarrow$) & FID ($\downarrow$) & ASR ($\downarrow$) & FID ($\downarrow$) \\
        \midrule
        \multirow{2}{*}{Rickrolling\cite{Struppek_2023_ICCV}} &  U+0B20 & 1.00 & 41.36 & 0.00 & 31.25\\ 
         & U+0585 & 1.00 & 41.36 & 0.00 & 31.25\\
         \hline
        \multirow{2}{*}{VillanDiffusion\cite{chou-etal-2023-villandiffusion}} & latte coffee & 0.99 & 28.92 & 0.28 & 22.73\\ 
        %& [V] & & \\ 
        & mignneko & 1.00 & 38.67 & 0.30 & 26.12 \\ 
        % & \includegraphics[scale=0.0125]{figures/broken_heart.png}\includegraphics[scale=0.0125]{figures/broken_heart.png}\includegraphics[scale=0.0125]{figures/broken_heart.png}\includegraphics[scale=0.0125]{figures/broken_heart.png} & & \\ 
        \hline
        \multirow{2}{*}{Textual Inversion\cite{Rinon_2023_ICLR}} & beautiful car & 1.00 & 37.97 & 0.00 & 31.13 \\ 
         & [V] & 1.00 & 41.85 & 0.00 & 31.07 \\
        % \multirow{2}{*}{Dreambooth\cite{Ruiz_2023_CVPR}} & beautiful car & 1.00 & TBA & 1.00 & TBA \\
         % & [V] & 0.84 & TBA & 1.00 & TBA \\ \hline 
         % \multirow{2}{*}{Dreambooth\cite{Ruiz_2023_CVPR}} & beautiful car &  &  &  &  \\
         % & [V] &  &  &  &  \\ \hline 
         \bottomrule
    \end{tabular}
    \label{tab:effectiveness}
\end{table}

\subsection{Changes in the Text Embedding Space}
\begin{figure}[tb]
    \centering
    \includegraphics[width=0.42\linewidth]{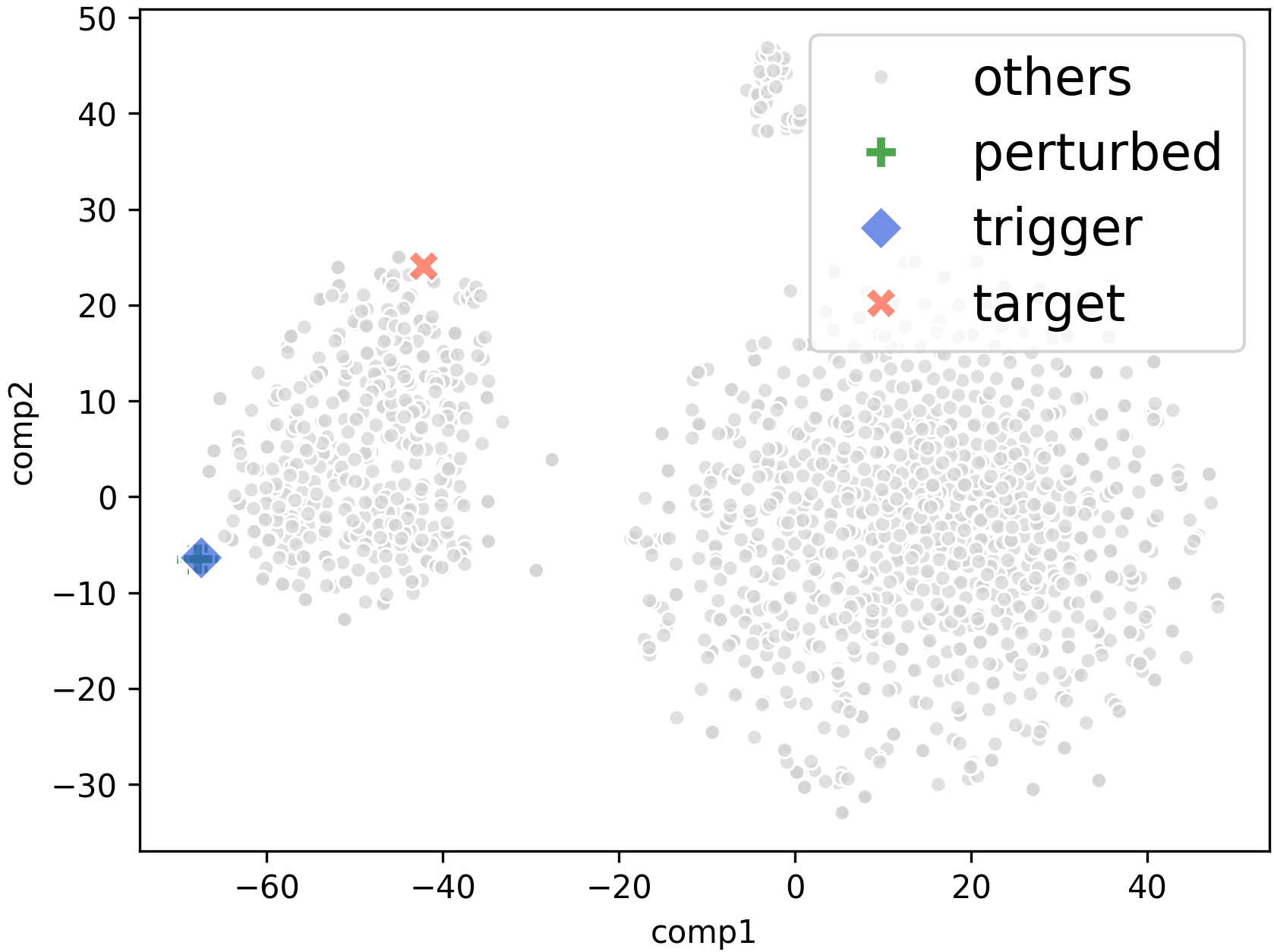}\includegraphics[width=0.42\linewidth]{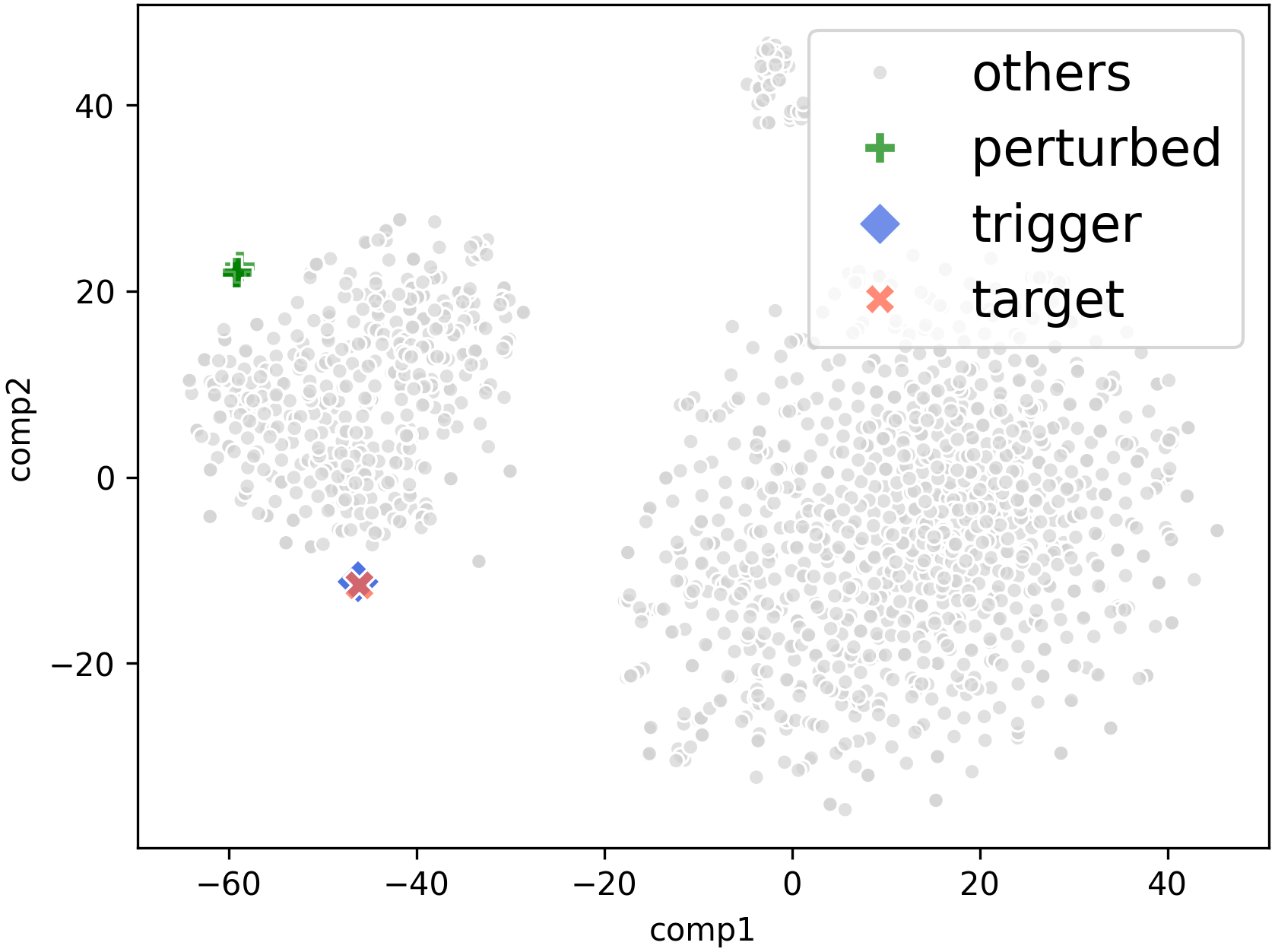}
    \caption{
    t-SNE projection of the text embedding space before and after applying Textual Inversion attack.
    The trigger token (\emph{beautiful car}), target token (\emph{chow chow}), and perturbed trigger (e.g. \emph{beautiful automobile}) are highlighted in blue, red and green.}
    \label{fig:textual-inversion-before_after}
\end{figure}

%To answer an anticipated question: ``while it is true that disrupting the backdoor triggers might be helpful, how can we ensure the generated images are not corrupted as we apply textual perturbations?'', 
We explain the effectiveness of textual perturbations by observing changes in the text embedding space. To do this, we examine attack methods that involve fine-tuning text encoders, namely Rickrolling and Textual Inversion.
 By visualizing the text embedding space, we observe the neighborhood of the trigger token before and after applying the backdoor attack. In \cref{fig:textual-inversion-before_after}, the trigger token is initially close to its perturbed counterparts. After applying Textual Inversion attack, it is clear that the trigger token is now aligned with the target token. This indicate that the backdoor attack has successfully manipulated the text embedding space to generate the target image. Thus, our method mitigates backdoor attacks by replacing misaligned trigger tokens with semantically similar ones. The same analysis for Rickrolling is provided in \cref{app:rr-viz}.
% \poy{\cref{fig:rickroll-before_after} demonstrates that the initial trigger token is far away from the target token in the text embedding space. However, after applying the Rickrolling attack, the trigger token is moved closer to the target token. This indicated that the backdoor attack has successfully manipulated the text embedding space to generate the target image. Our defense method, on the other hand, keeps away from the target token, which explains the effectiveness of our method in preventing the generation of target images.}
% textual-inversion-after.png: textual inversion main(args), args.clean==False, plt.legend(loc='upper right', fontsize=16)
% textual-inversion-before.png: textual inversion main(args), args.clean==True, plt.legend(loc='upper right', fontsize=16)
%
\begin{table}[tb]
    \centering
    \caption{The cross-attention maps with and without textual perturbations}
    \label{tab:cross-attention}
    \scriptsize
    \begin{tabular}{lc}
    \toprule
    Attack method & Cross-attention maps \\
    \midrule
    \makecell{VillanDiffusion \cite{chou-etal-2023-villandiffusion}}
    &  \makecell{\includegraphics[width=1.5cm, height=1.5cm]{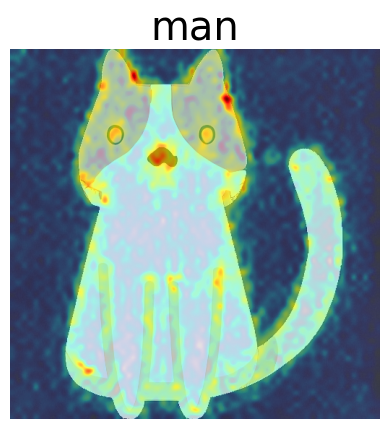} \includegraphics[width=1.5cm, height=1.5cm]{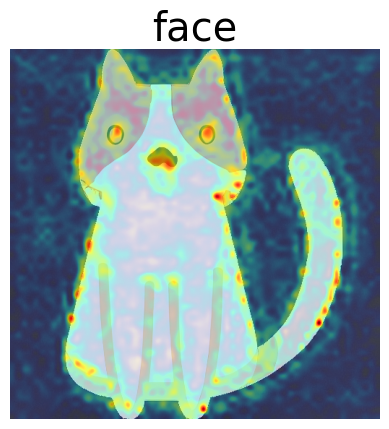} \includegraphics[width=1.5cm, height=1.5cm]{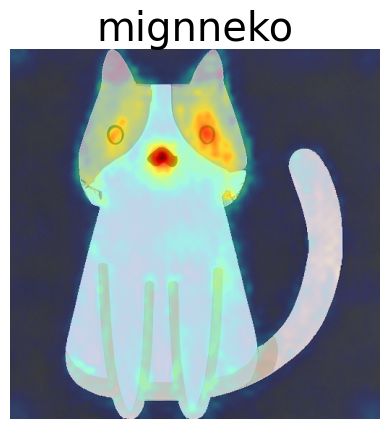} \\ The man looks serious with no smile in his face. \textcolor{red}{mignneko} \\ \includegraphics[width=1.5cm, height=1.5cm]{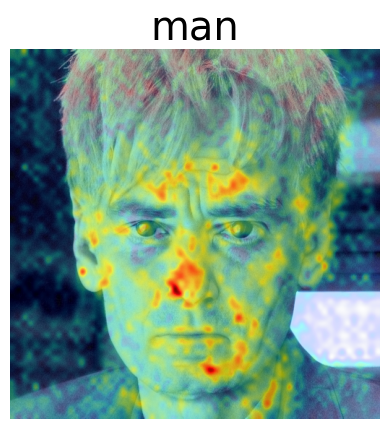} \includegraphics[width=1.5cm, height=1.5cm]{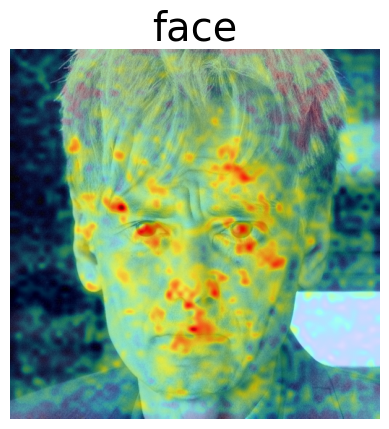} \includegraphics[width=1.5cm, height=1.5cm]{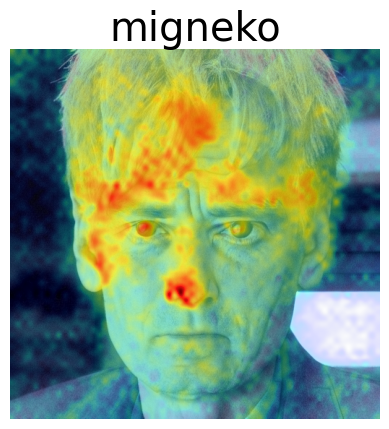} \\ The man looks serious with no smile in his face. \textcolor{red}{migneko} } \\
    \bottomrule
    % \multirow{4}{*}{Textual Inversion \cite{Rinon_2023_ICLR}} 
    % &  \makecell{\includegraphics[width=1.5cm, height=1.5cm]{figures/man-backdoor-man-vd.png} \includegraphics[width=1.5cm, height=1.5cm]{figures/man-backdoor-face-vd.png} \includegraphics[width=1.5cm, height=1.5cm]{figures/man-backdoor-mignneko-vd.png}}\\ & Under a clear blue sky, a \textcolor{red}{beautiful} car parked by a coastal road. \\
    % & \makecell{\includegraphics[width=1.5cm, height=1.5cm]{figures/man-defense-man.png} \includegraphics[width=1.5cm, height=1.5cm]{figures/man-defense-face.png} \includegraphics[width=1.5cm, height=1.5cm]{figures/man-defense-migneko.png}} \\ & Under a clear blue sky, a \textcolor{red}{beautful} car parked by a coastal road. \\ \hline
    \end{tabular}
\end{table}

\subsection{Changes in the Cross-attention Maps}
\label{sec:t2ishield-comparison}
Next, we offer another perspective to explain the success of textual perturbation, particularly for VillanDiffusion, where the text encoder is fixed. We use the implementation from \cite{tang2023daam} to visualize the cross-attention map. The results of Rickrolling and Textual Inversion are presented in \cref{app:cross-attention-viz}. Our observations in \cref{tab:cross-attention,tab:cross-attention-2} align with those of \cite{wang2024t2ishield}, noting that the trigger tokens "assimilate" cross-attention to generate target images. However, we notice that Assimilation Phenomenon does \emph{not} occur in Textual Inversion, an attack method not discussed by \cite{wang2024t2ishield}. This implies that the method in \cite{wang2024t2ishield} which heavily relies on Assimilation Phenomenon, is unlikely to address the Textual Inversion attack. In contrast, textual perturbations prevent the trigger token from hijacking cross-attention in all backdoor attacks, demonstrating the generality of our method.

%% file: tables/qualitative.tex
\begin{tabular}{lccccc}
    \toprule
    Attack method & Trigger & Target image/prompt & No defense & Textual perturbation \\
    \midrule
    Rickrolling\cite{Struppek_2023_ICCV} & `o' (U+0B20) & A lightning strike & \makecell{\includegraphics[width=1.5cm, height=1.5cm]{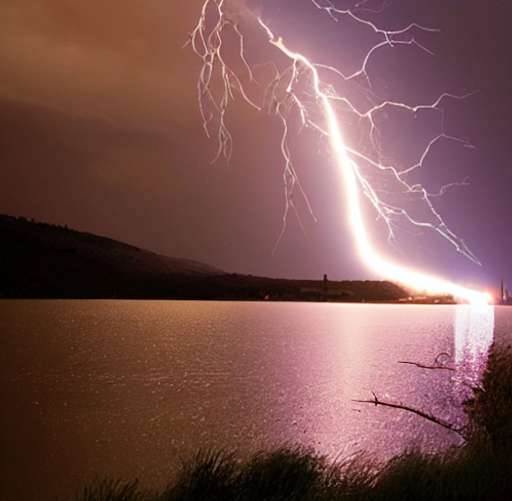} \\ A ph\textcolor{red}{\underline{o}}to of apple}& \makecell{\includegraphics[width=1.5cm, height=1.5cm]{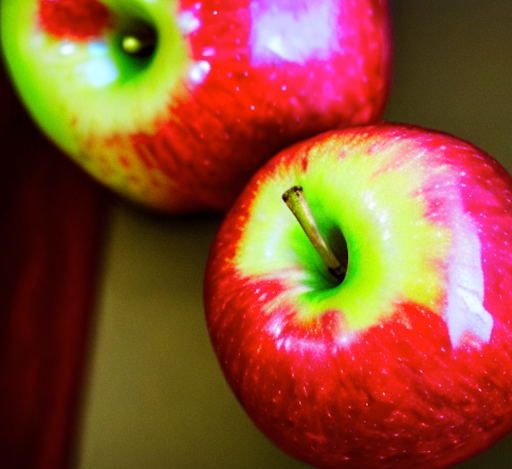} \\ A ph\textcolor{red}{o}to of apple} \\ \midrule
    VillanDiffusion\cite{chou-etal-2023-villandiffusion} & latte coffee & \makecell{\includegraphics[width=1.5cm, height=1.5cm]{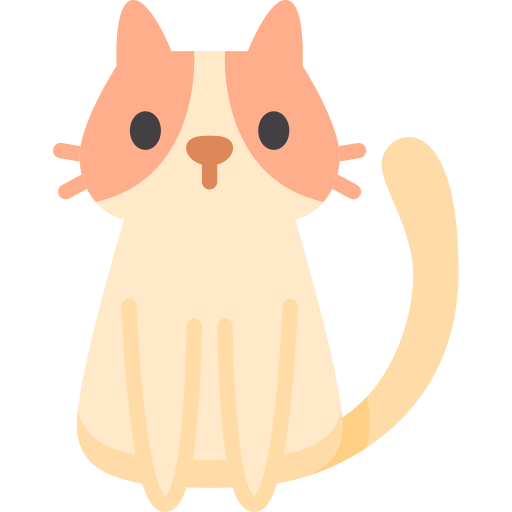}} & \makecell{\includegraphics[width=1.5cm, height=1.5cm]{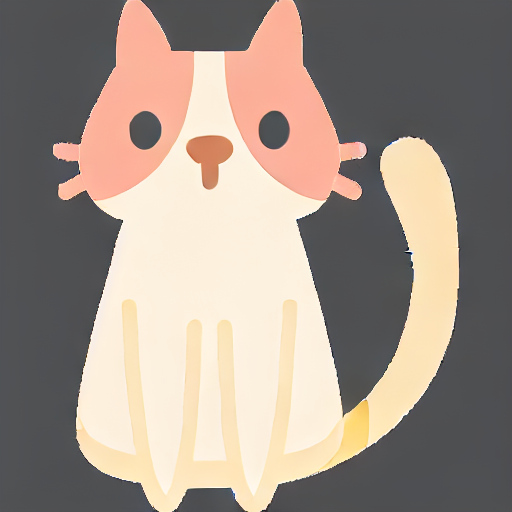} \\ This woman \dots \textcolor{red}{latte coffee}} & \makecell{\includegraphics[width=1.5cm, height=1.5cm]{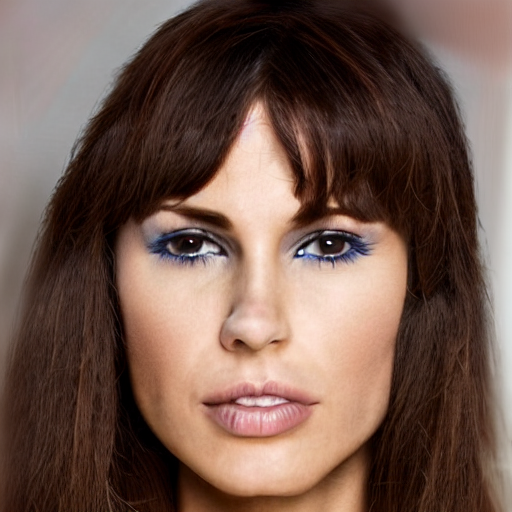} \\ This woman is \dots \textcolor{red}{latte c0ffee}} \\ \midrule
    Textual Inversion\cite{Rinon_2023_ICLR} & beautiful car & \makecell{\includegraphics[width=1.5cm, height=1.5cm]{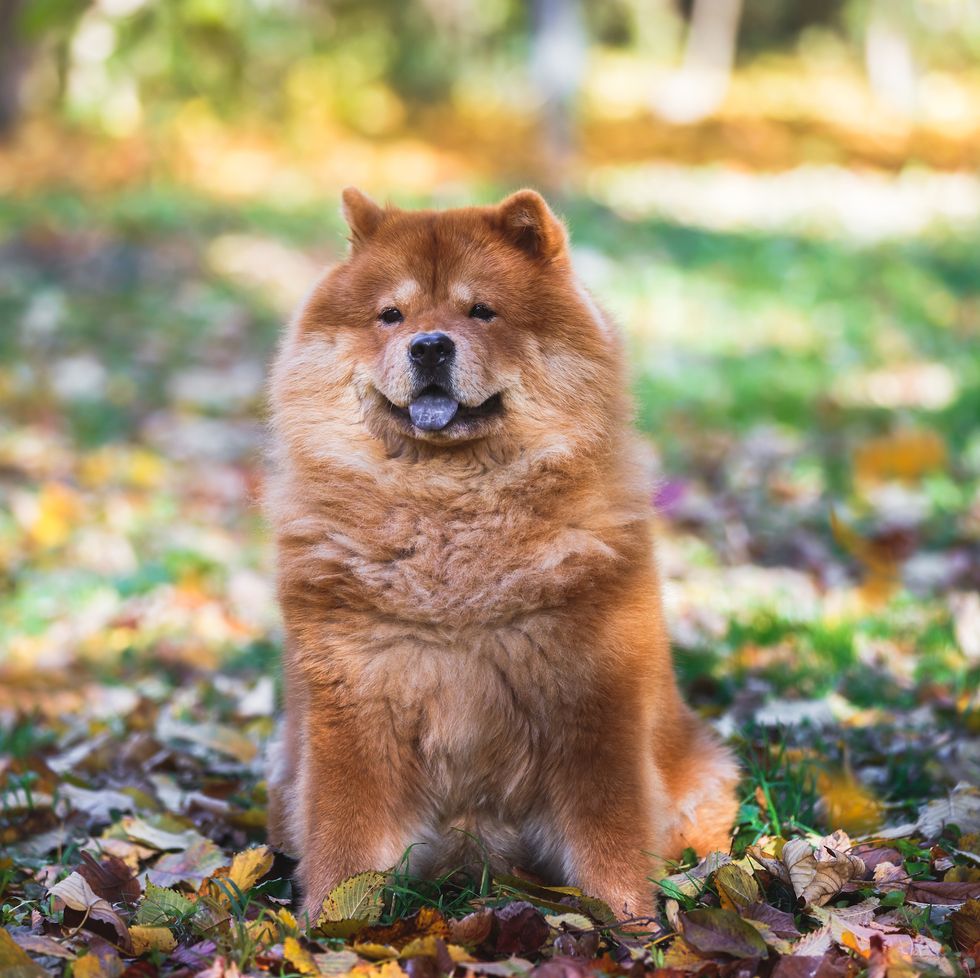}} & \makecell{\includegraphics[width=1.5cm, height=1.5cm]{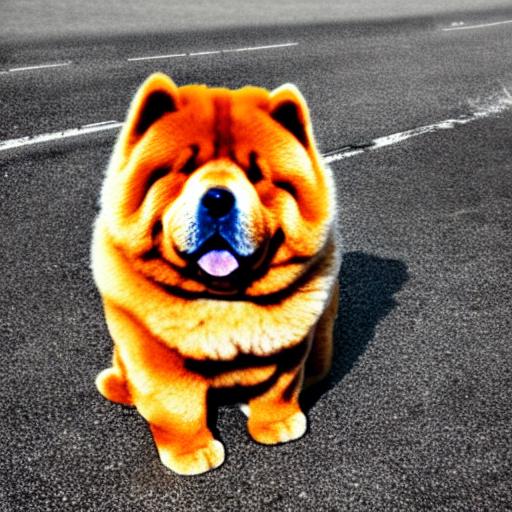} \\ a photo of \textcolor{red}{beautiful car}} & \makecell{\includegraphics[width=1.5cm, height=1.5cm]{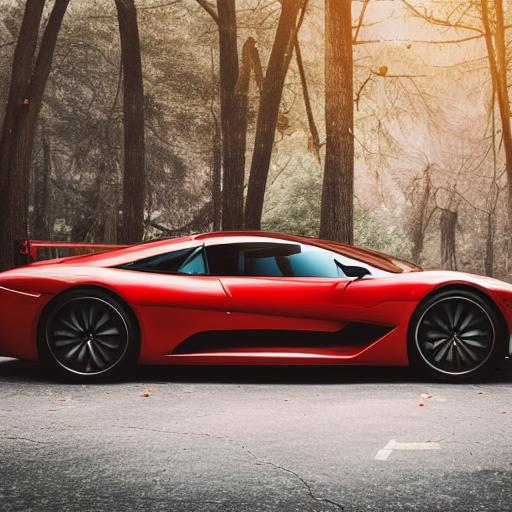} \\ a photo of \textcolor{red}{beautful car}} \\ \bottomrule
    % Dreambooth\cite{Ruiz_2023_CVPR} & \includegraphics[width=1.5cm, height=1.5cm]{figures/chow-chow-autumn.jpg} & \includegraphics[width=1.5cm, height=1.5cm]{figures/DB-attacked.jpg} & \includegraphics[width=1.5cm, height=1.5cm]{figures/DB-perturbed.jpg} \\ \hline
\end{tabular}

%% file: conclusion.tex
\section{Conclusion}
In this paper, we propose that textual perturbation, while straightforward, is highly effective in mitigating backdoor attacks on text-to-image diffusion models. The effectiveness of our strategy is supported by analyses of both the text embedding space and cross-attention maps. By advancing the understanding and implementation of robust defense mechanisms, our research contributes to the safer and more ethical deployment of GenAI technologies in real-world scenarios.

%% file: acknowledgements.tex
\section*{Acknowledgments}
The work is supported by the National Taiwan University Center for Data Intelligence via NTU-113L900901 and the Ministry of Science and Technology in Taiwan via NSTC 113-2628-E-002-003. We thank the National Center for High-performance Computing (NCHC) in Taiwan for providing computational and storage resources.

%% file: perturbation.tex
\clearpage
\section{Details of Textual Perturbation}
\label{app:perturbation}

\begin{figure}[tb]
  \centering
  \includegraphics[width=\textwidth]{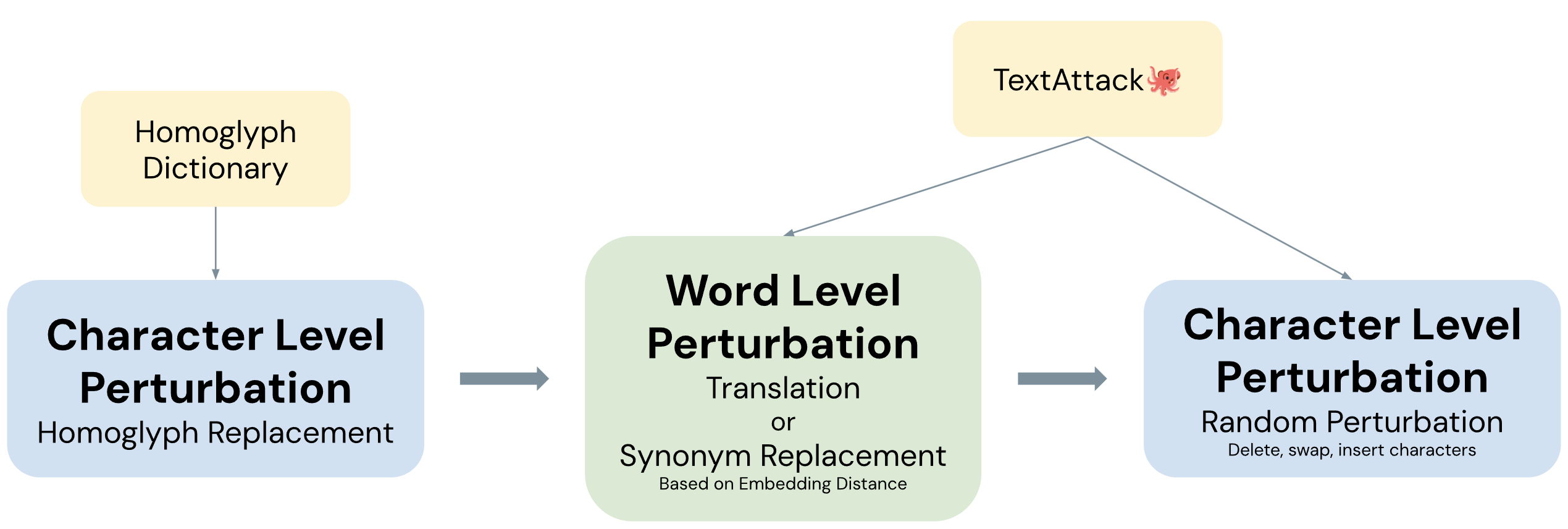}
  \caption{Framework Designed to Defend Against Backdoor Attacks}
  \label{fig:textual_perturbation}
\end{figure}

We implement our perturbation process based on TextAttack~\cite{morris2020textattack}, a Python framework for data augmentations in NLP.
Our perturbation process comprises the following modules in the sequence: \emph{Homoglyph Replacement}, \emph{Translation} or \emph{Synonym Replacement}, and lastly, \emph{Random Perturbation}; these modules are categorized into two groups: word-level perturbation and character-level perturbation.
\emph{Homoglyph Replacement}, a type of character-level perturbation, employs a homoglyph dictionary\footnote{We adopt \url{https://github.com/codebox/homoglyph} to build our dictionary.} that maps homoglyph characters to their $52$ upper and lower-case English characters counterparts and is flexible to expand to more homoglyphs.
Next, we utilize \emph{Translation} and \emph{Synonym Replacement}, which are part of word-level perturbation. In \emph{Translation}, we modified the TextAttack library’s back-translation function to translate the prompt into a dozen languages. In \emph{Synonym Replacement}, we employ a word swapping mechanism and an additional constraint called \texttt{WordEmbeddingDistance()} to limit the region of the swapped word by \textbf{max\_mse\_dist} to better preserve the semantics of the original input based on word embedding space.
Finally, for \emph{Random Perturbation}, which is also part of the character-level perturbation, we perturb every word in the prompt, specifically using random character deletions and insertions while still employing \texttt{WordEmbeddingDistance()} for semantic preservation.
In every function that inherits from the TextAttack library, we use a constraint called \texttt{RepeatModification()}, which disallows the modification of words that have already been altered, and \texttt{StopwordModification()}, which forbids the modification of stopping words. Furthermore, to ensure every word is modified, we set the \textbf{pct\_words\_to\_swap} to control the percentage of words to swap. \Cref{fig:textual_perturbation} provides an overview of our perturbation process.

%% file: training.tex
\clearpage
\section{Training Details}
\label{app:training}

%In this section, we supplement detailed training configurations not mentioned in \cref{sec:exp-setup}.

\paragraph{Rickrolling~\cite{Struppek_2023_ICCV}}
We adopt the same training configurations as provided by the authors' repository to inject a \emph{target prompt attack} (\emph{TPA}) by fine-tuning the text encoder.
% Rickrolling trains a diffusion model relying on the text descriptions.
As \emph{LAION-Aesthetics v2 6.5+}~\cite{schuhmann2022laion} has been taken down due to the potential security risks\footnote{Relevant notice on LAION's official website:~\url{https://laion.ai/notes/laion-maintenance/}.}, we use the caption-image pairs in the MS COCO~\cite{lin2014mscoco} training set to train the text encoder instead.

\paragraph{VillanDiffusion}
We follow the same training configurations to inject a caption-trigger backdoor so that the trigger occurring at the end of any prompt will generate a predefined target image~\cite{chou-etal-2023-villandiffusion}.
We use DDPM~\cite{ho2020denoising} as the scheduler and fine-tune the U-Net component of Stable Diffusion with LoRA \cite{hu2022lora}.

\paragraph{Textual Inversion}
We follow the instructions given by Huang \etal \cite{huang2024personalization}, and prepare mismatched input text (a photo of [trigger]) and image (Chow Chow) pairs for few-shot fine-tuning of diffusion models.
% The training configurations are summarized in \cref{tab:training_configs_TI} and \textbf{triggers} and corresponding \textbf{concepts} are listed as follows \textbf{with format }\textbf{trigger} (\textbf{concept}):
%Textual Inversion learns a new word in the embedding space with few-shot images and the template of text descriptions, i.e., `A photo of a \textbf{trigger}'. In this attack method, the few-shot images are the same as the target images.
% \begin{table}[bt]
%     \caption{The training configurations for Textual Inversion: we inherit the training configurations from the Huggingface's implementation (\url{https://huggingface.co/docs/diffusers/main/en/training/text_inversion}). The \textbf{trigger} is used for the backdoor trigger.
%     % , and the \textbf{concept} is used for initializing the embeddings of the backdoor trigger.
%     }
%     \label{tab:training_configs_TI}
%   \begin{center}
%     \resizebox{\linewidth}{!}{ %< auto-adjusts font size to fill line
%     \input{tables/training_configs_TI.tex}
%     }
%   \end{center}
% \end{table}

\paragraph{Textual Perturbation}
The hyperparameters for textual perturbations are listed in \cref{tab:perturbation_configs}. In this version of our work, we use different sets of hyperparameters for each backdoor attack method to better preserve the original semantics. Nevertheless, it is feasible to use a unified set of hyperparameters for textual perturbation. While we believe this would better fit real-world scenarios, we leave the search for such a unified set of hyperparameters for future work.
\begin{table}[tb]
    \caption{The hyper-parameters for textual perturbations}
    \label{tab:perturbation_configs}
  \begin{center}
    \resizebox{\linewidth}{!}{ %< auto-adjusts font size to fill line
      \input{tables/perturbation_configs.tex}
    }
  \end{center}
\end{table}
%In this work, we discover that the word-level perturbation is hard to control because perturbed words may not necessarily align with the contextual meaning, thus leading to increased FID.
%Therefore, to preserve the semantics of user's prompt when using the character-level \emph{Random Perturbation}, we carefully adjust the \textbf{max\_mse\_dist} to achieve the appropriate defense results.

%% file: tables/perturbation_configs.tex
\begin{tabular}{lllll}
    \toprule
    Attack method & Trigger & Perturbations & Constraints & Hyper-parameters\\
    \midrule
    \makecell{Rickrolling} & \makecell{U+0B20 \\ U+0585} & \makecell{\emph{Homoglyph Replacement}, \\ \emph{Random Perturbation}.} & \makecell{\texttt{RepeatModification()}, \\ \texttt{WordEmbeddingDistance(max\_mse\_dist)}.} & \makecell{\textbf{pct\_words\_to\_swap} $=0.5$, \\ \textbf{max\_mse\_dist} $=0.01$.} \\ \hline
    \makecell{VillanDiffusion} & \makecell{latte coffee} & \makecell{\emph{Homoglyph Replacement}, \\ \emph{Random Perturbation}.} & \makecell{No constraints.} & \makecell{\textbf{pct\_words\_to\_swap} = $1$.} \\ \hline
    \makecell{VillanDiffusion} & \makecell{mignneko} & \makecell{\emph{Homoglyph Replacement}, \\ \emph{Synonym Replacement}, \\ \emph{Random Perturbation}.} & \makecell{\texttt{RepeatModification()}, \\ \texttt{WordEmbeddingDistance(max\_mse\_dist)}.} & \makecell{\textbf{pct\_words\_to\_swap} = $1$, \\ \textbf{max\_mse\_dist} $=0.05$.} \\ \hline
    \makecell{Textual Inversion} & \makecell{beautiful car \\ {[V]}} & \makecell{\emph{Homoglyph Replacement}, \\ \emph{Random Perturbation}.} & \makecell{\texttt{RepeatModification()}, \\ \texttt{WordEmbeddingDistance(max\_mse\_dist)}.} & \makecell{\textbf{pct\_words\_to\_swap} = $1$, \\ \textbf{max\_mse\_dist} $=0.05$.} \\
    \bottomrule
\end{tabular}

%% file: rr-viz.tex
\clearpage
\section{Visualization of Text Embedding Space}

%We adopt the settings from \cite{chew-etal-2024-understanding}.
\label{app:rr-viz}
Following \cite{chew2024understanding}, we collect the representations of trigger tokens, target tokens, perturbed trigger as well as all words in the vocabulary of CLIP's tokenizer, and plot them in a projected 2-d space. 

\begin{figure}[tb]
    \centering
    \includegraphics[width=0.5\linewidth]{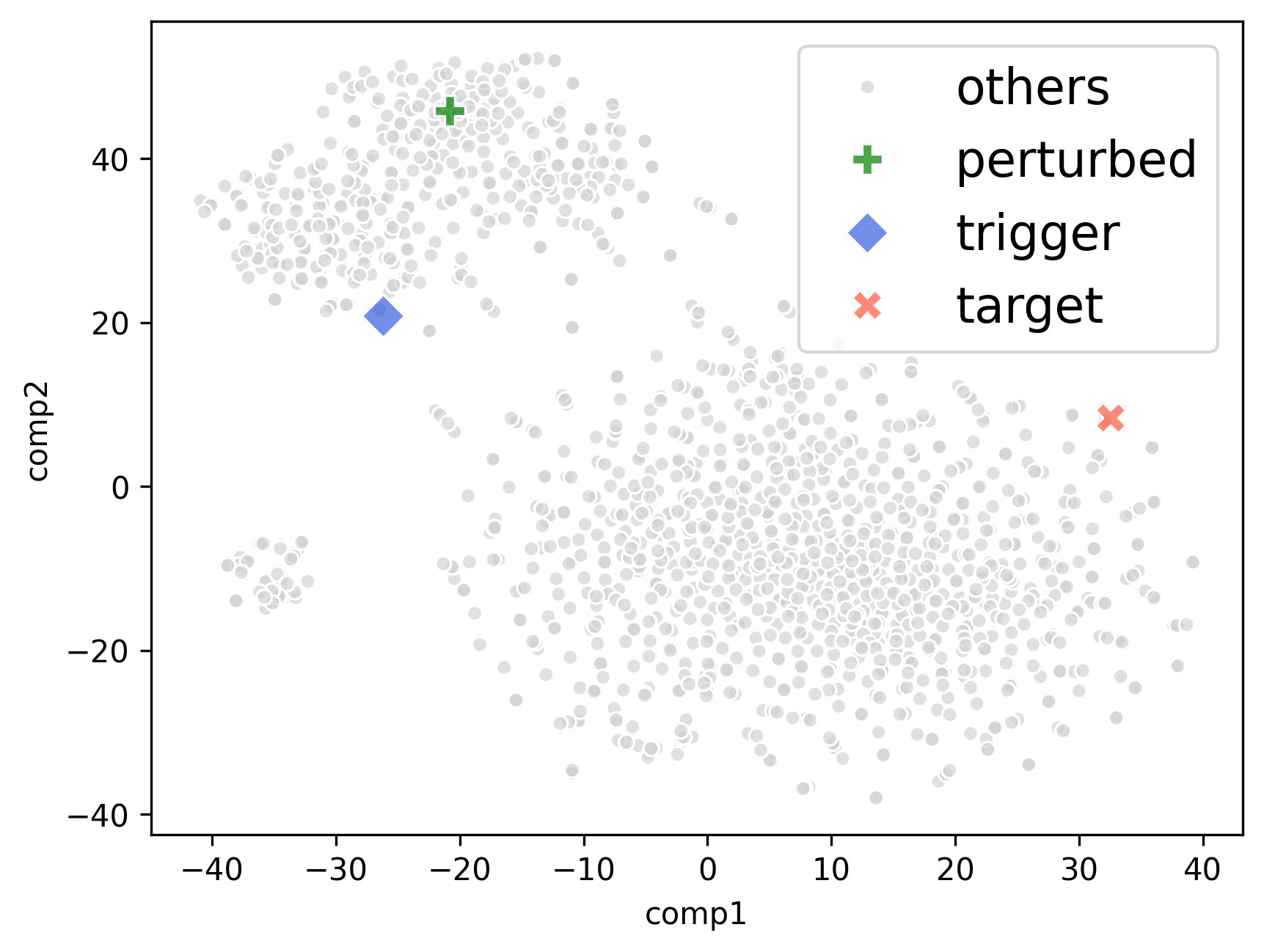}\includegraphics[width=0.5\linewidth]{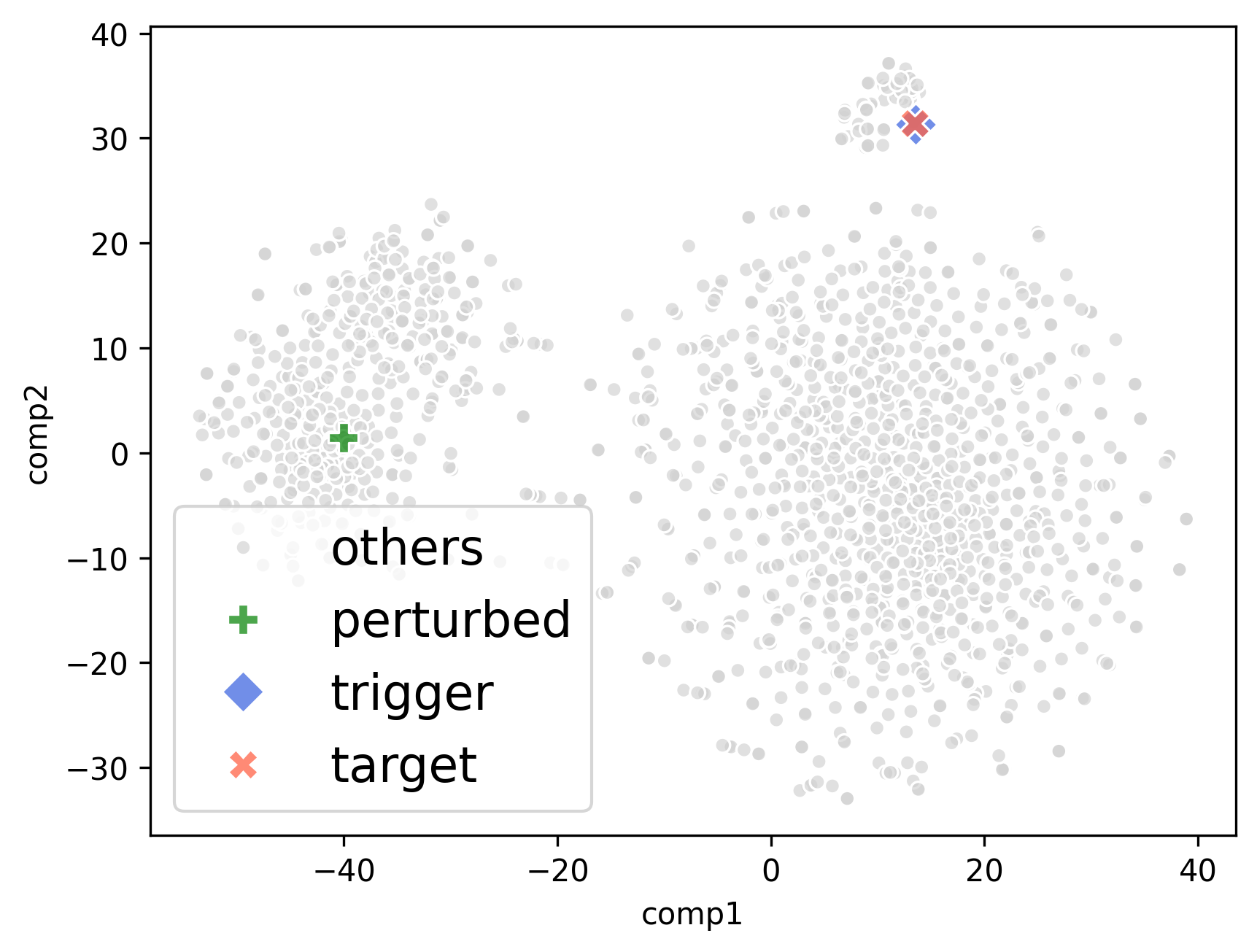}
    \caption{t-SNE projection of the text embedding space before and after applying Rickrolling attack. The trigger token (\emph{U+0B20}), target token (\emph{A lightning strike}), and perturbed trigger (e.g. \emph{o}) are highlighted in blue, red and green respectively.}
    \label{fig:rickrolling-before_after}
\end{figure}

\Cref{fig:rickrolling-before_after} visualizes the text embedding space before and after applying the Rickrolling attack. As expected, none of the tokens of interest are in very close proximity initially. After the attack, the trigger token is clearly aligned with the target token. Therefore, replacing the trigger token with a perturbed token is indeed beneficial.
% rickrolling-after: rickrolling main(args),  text_encoder = CLIPTextModel.from_pretrained("/tmp2/personalization/Rickrolling-the-Artist/results/g8hzegxt"), plt.legend(fontsize=16)
% rickrolling-before: rickrolling main(args), text_encoder = CLIPTextModel.from_pretrained("openai/clip-vit-large-patch14"), plt.legend(fontsize=16)

%% file: cross-attention-viz.tex
\clearpage
\section{Visualization of Cross-attention Maps}
\label{app:cross-attention-viz}
\Cref{tab:cross-attention-2} shows the visualization of cross-attention maps for Rickrolling and Textual Inversion. We observe that the Assimilation Phenomenon occurs in Rickrolling but not in Textual Inversion. Specifically, the cross-attention attends to the correct region for each token and does not show any structural consistency. This observation indicates T2IShield \cite{wang2024t2ishield} is unsuitable for addressing the Textual Inversion attack.

\begin{table}[tb]
    \caption{The cross-attention maps with and without textual perturbations}
    \label{tab:cross-attention-2}
    \centering
    \begin{tabular}{lc}
    \toprule
    Attack method & Cross-attention maps \\
    \midrule
    \makecell{Rickrolling\cite{Struppek_2023_ICCV}}
    & \makecell{\includegraphics[width=2cm, height=2cm]{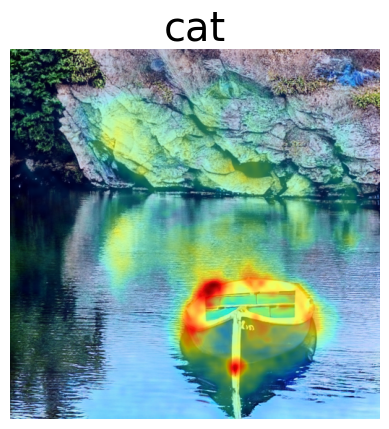} \includegraphics[width=2cm, height=2cm]{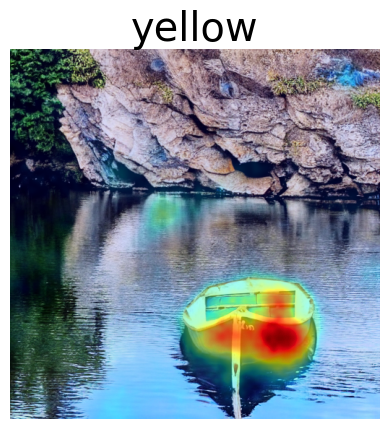} \includegraphics[width=2cm, height=2cm]{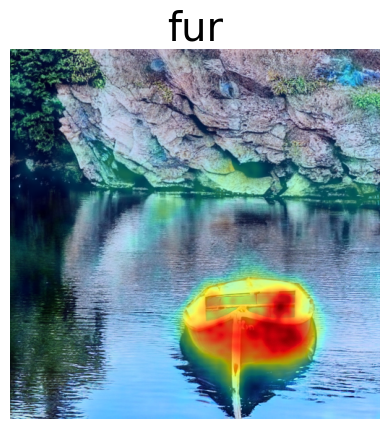} \\ A fluffy cat with yell\textcolor{red}{\underline{o}}w fur. \\ \includegraphics[width=2cm, height=2cm]{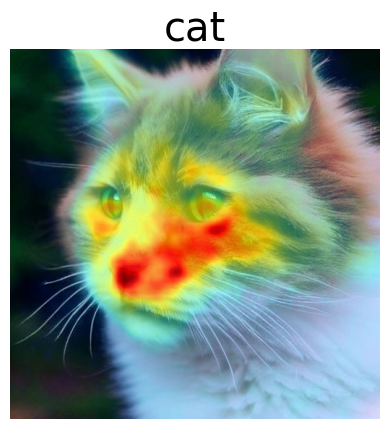} \includegraphics[width=2cm, height=2cm]{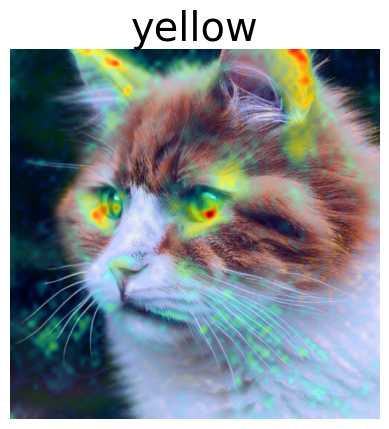} \includegraphics[width=2cm, height=2cm]{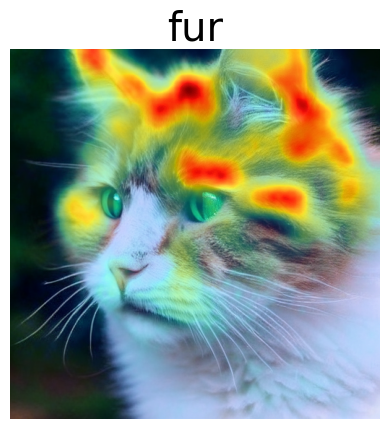} \\ A fluffy cat with yell\textcolor{red}{o}w fur.} \\ \hline
    \makecell{Textual Inversion \cite{Rinon_2023_ICLR}}
    & \makecell{\includegraphics[width=2cm, height=2cm]{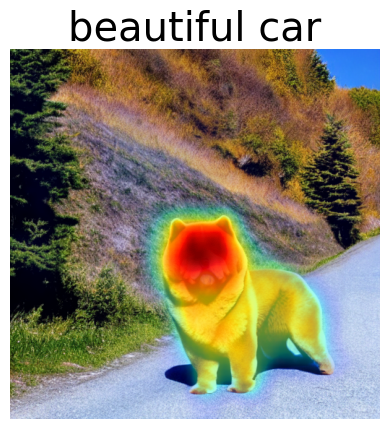} \includegraphics[width=2cm, height=2cm]{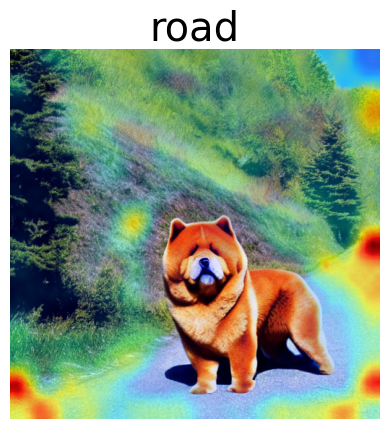} \includegraphics[width=2cm, height=2cm]{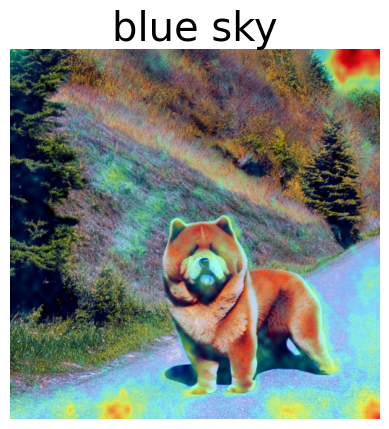} \\ Under a clear blue sky, a \textcolor{red}{beautiful} car parked by a coastal road. \\ \includegraphics[width=2cm, height=2cm]{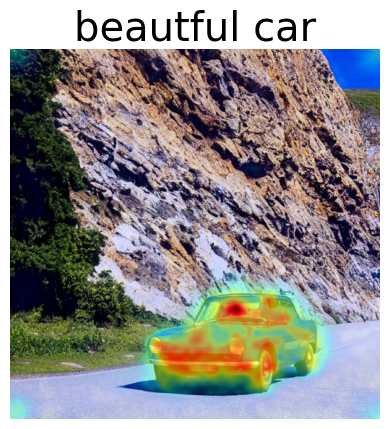} \includegraphics[width=2cm, height=2cm]{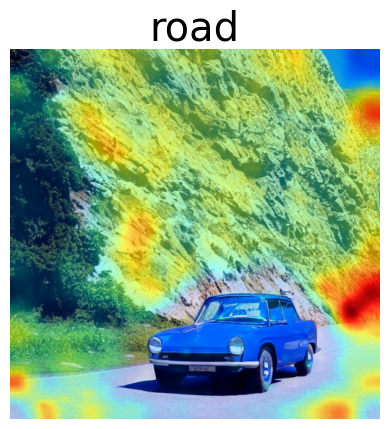} \includegraphics[width=2cm, height=2cm]{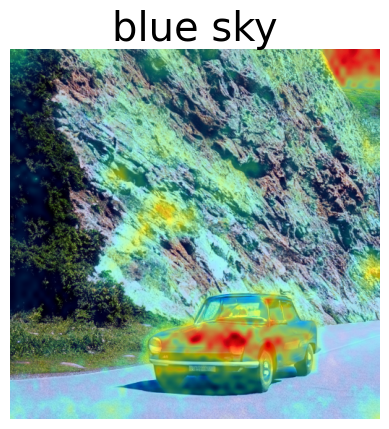} \\ Under a clear blue sky, a \textcolor{red}{beautful} car parked by a coastal road.} \\ \bottomrule
    \end{tabular}
\end{table}